\documentclass[10pt,twocolumn,letterpaper]{article}

\usepackage{iccv}
\usepackage{times}
\usepackage{epsfig}
\usepackage{graphicx}
\usepackage{amsmath, bm}
\usepackage{amssymb}

\usepackage{enumitem}
\usepackage{multirow}
\usepackage{siunitx}
\usepackage{bbold}
\usepackage{bm}

\usepackage[breaklinks=true,bookmarks=false]{hyperref}

\iccvfinalcopy 

\def\iccvPaperID{3280} 

\ificcvfinal\fi
\begin{document}

\title{Balanced Datasets Are Not Enough: \\ Estimating and Mitigating Gender Bias in Deep Image Representations}

\author{Tianlu Wang$^1$, Jieyu Zhao$^2$,  Mark Yatskar$^3$, Kai-Wei Chang$^2$, Vicente Ordonez$^1$\\
$^1$University of Virginia, $^2$University of California Los Angeles, \\$^3$Allen Institute for Artificial Intelligence\\
{\tt\small tianlu@virginia.edu, jyzhao@cs.ucla.edu, marky@allenai.org,}\\{ \tt\small kwchang@cs.ucla.edu, vicente@virginia.edu}
}

\maketitle
\ificcvfinal\thispagestyle{empty}\fi

\begin{abstract}
In this work, we present a framework to measure and mitigate intrinsic biases with respect to protected variables --such as gender-- in visual recognition tasks. 
We show that 
trained models significantly amplify the association of target labels with gender beyond what one would expect from biased datasets.
Surprisingly, we show that even when datasets are balanced such that each label co-occurs equally with each gender, learned models amplify the association between labels and gender, as much as if data had not been balanced! 
To mitigate this, we adopt an adversarial approach to remove unwanted features corresponding to protected variables from intermediate representations in a deep neural network -- and provide a detailed analysis of its effectiveness. 
Experiments on two datasets: the COCO dataset (objects), and the imSitu dataset (actions), show reductions in gender bias amplification while maintaining most of the accuracy of the original models. 

\end{abstract}

\section{Introduction}
While visual recognition systems have made great progress toward practical applications, they are also sensitive to spurious correlations and often depend on these erroneous associations.
When such systems are used on images containing people, they risk amplifying societal stereotypes by over associating protected attributes such as gender, race or age with target predictions, such as object or action labels. 
Known negative outcomes have included representation harms (e.g., male software engineers are being over-represented in image search results~\cite{kay2015unequal}), harms of opportunity, (e.g., facial recognition is not as effective for people with different skin tones~\cite{pmlr-v81-buolamwini18a}), to life-threatening situations (e.g., recognition rates of pedestrians in autonomous vehicles are not equally accurate for all groups of people~\cite{wilson2019predictive}).

\begin{figure}[t]
    \centering
      \includegraphics[width=0.48\textwidth]{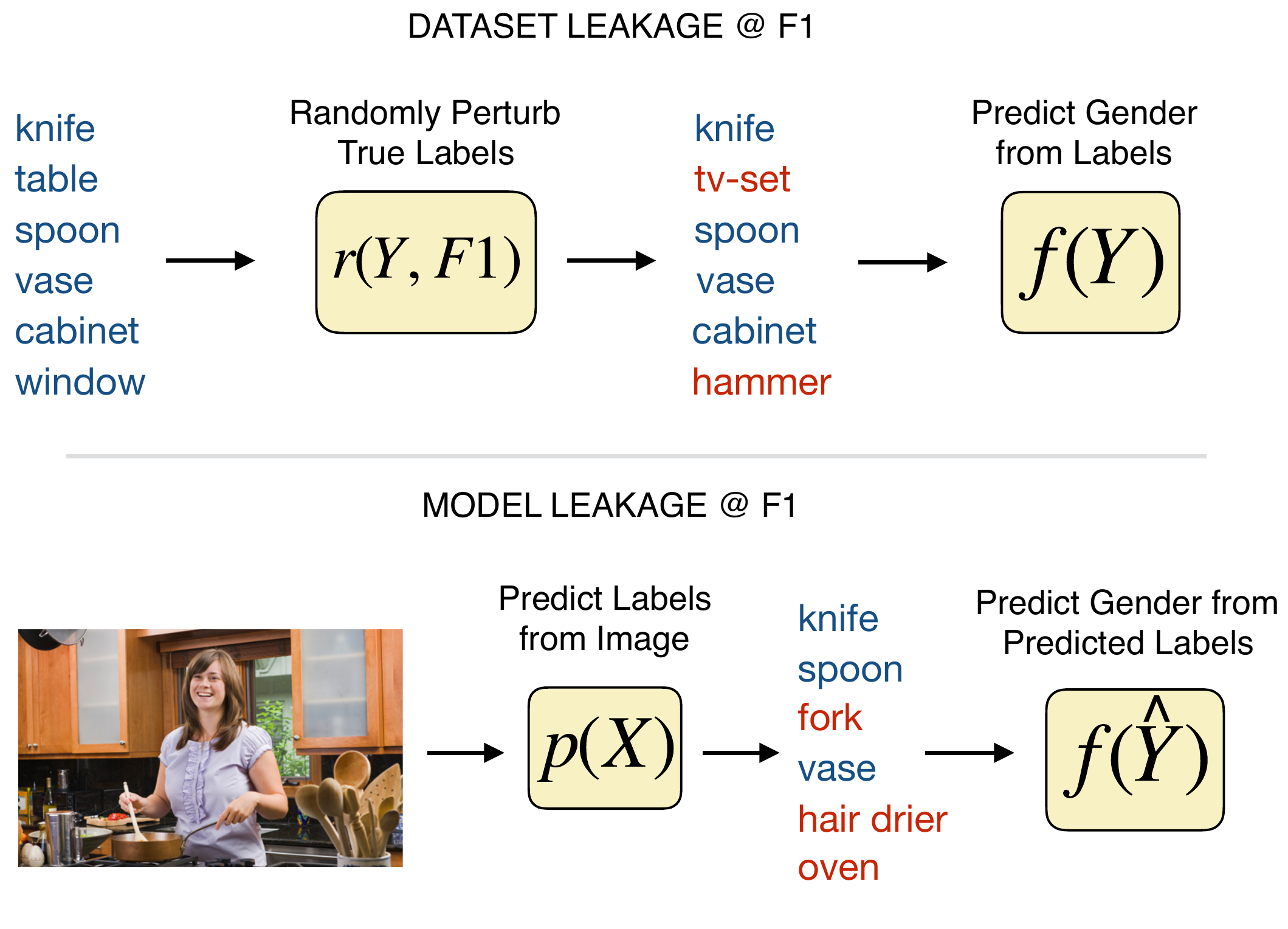}
    \caption{On the top we illustrate our newly introduced concept of \emph{Dataset Leakage} which measures the extent to which gender --or more generally a protected variable-- can be inferred from randomly perturbed ground-truth labels. On the bottom we illustrate our concept of \emph{Model Leakage} which measures the extent to which gender can be inferred from the outputs of a model. A model amplifies bias if model leakage exceeds dataset leakage. 
    }
    \vspace{-0.2in}
    \label{fig:lead}
\end{figure}

In this paper we study gender bias amplification: the effect that trained models exaggerate gender stereotypes that are present in the training data. 
We focus on the tasks of recognizing objects in the COCO dataset~\cite{lin2014microsoft} and actions in the imSitu dataset~\cite{yatskar2016}, where training resources exhibit gender skew and models trained on these datasets exhibit bias amplification~\cite{zhao2017men}.\footnote{For example women are represented as cooking twice as often as men in imSitu, but after models are trained and evaluated on similarly distributed data, they predict cooking for women three times as often as men.}
In an effort to more broadly characterize bias amplification, we generalize existing measures of bias amplification. 
Instead of measuring the similarity between training data and model prediction distributions, we compare the predictability of gender from ground truth labels ({\it dataset leakage}, Figure~\ref{fig:lead} on the top) and model predictions ({\it model leakage}, Figure~\ref{fig:lead} on the bottom).
Each of these measures is computed using a classifier that is trained to predict gender from either ground truth labels or models predictions.
We say a model exhibits bias amplification if it leaks more information about gender than a classifier of equivalent accuracy whose errors are only due to chance.

Our new leakage measures significantly expand the types of questions we can ask about bias amplification. 
While previously it was shown that models amplify bias when they are required to predict gender alongside target variables~\cite{zhao2017men}, our empirical findings indicate that when models are not trained to predict gender, they also amplify gender bias. 
Surprisingly, we find that if we additionally balance training data such that each gender co-occurs equally with each target variable, models amplify gender bias as much as in unbalanced data!
This strongly argues that naive attempts to control for protected attributes when collecting datasets will be ineffective in preventing bias amplification.

\begin{figure}[t]
    \centering
      \includegraphics[width=.45\textwidth]{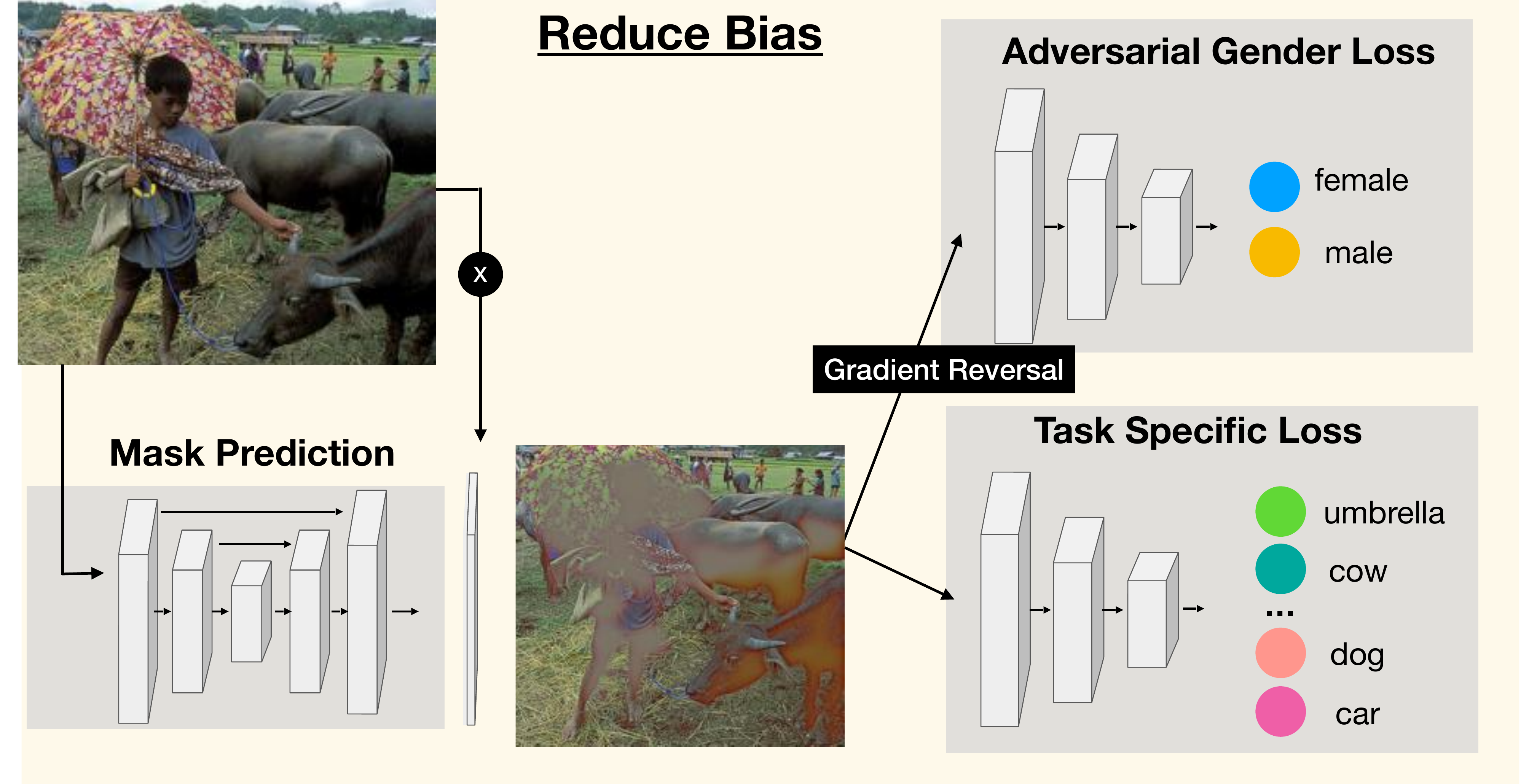}
    \caption{In our bias mitigation approach, we learn a task-specific model with an adversarial loss that removes features corresponding to a protected variable from an intermediate representation in the model -- here we illustrate our pipeline to visualize the removal of features in image space through an auto-encoder network.}
    \vspace{-0.2in}
    \label{fig:method}
\end{figure}

We posit that models amplify biases in the data balanced setting because there are many gender-correlated but unlabeled features that cannot be balanced directly. For example in a dataset with equal number of images showing men and women cooking, if \emph{children} are unlabeled but co-occur with the \emph{cooking} action, a model could associate the presence of children with \emph{cooking}. Since children co-occur with women more often than men across all images, a model could label women as \emph{cooking} more often than we expect from a balanced distribution, thus amplifying gender bias. 

\begin{figure}[t]
    \centering
      \includegraphics[width=.32\textwidth]{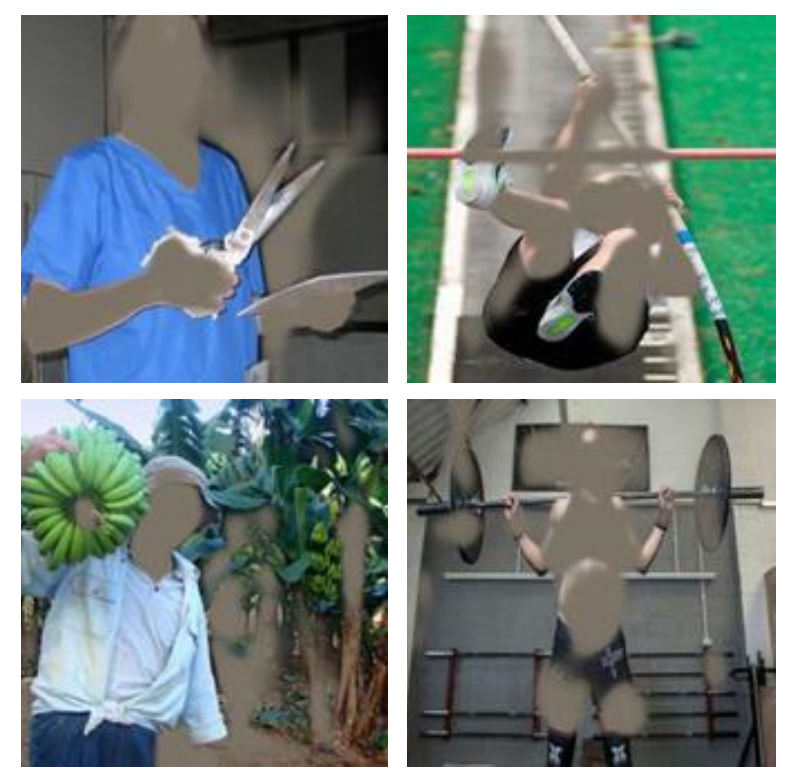}
    \caption{Images after adversarial removal of gender when applied to the image space. The objective was to preserve information about objects and verbs, e.g. scissors, banana (COCO) or vaulting, lifting (imSitu) while removing gender correlated features.}
    \vspace{-0.2in}
    \label{fig:method_lead}
\end{figure}

To mitigate such unlabeled spurious correlations, we adopt an adversarial debiasing approach~\cite{xie2017controllable,beutel2017data,zhang2018mitigating,elazar2018adversarial}.
Our goal is to preserve as much task specific information as possible while eliminating gender cues either directly in the image or intermediate convolutional representations used for classification.
As seen in Figure~\ref{fig:method}, models are trained adversarially to trade off a task-specific loss while trying to create a representation from which it is not possible to predict gender.
For example, in Figure~\ref{fig:method_lead} in the bottom right image, our method is able to hide regions that indicate the gender of the main entity while leaving enough information to determine that {\it she} is weight lifting.

Evaluation of our adversarial debiased models show that they are able to make significantly better trade-offs between task accuracy and bias amplification than other methods.
We consider strong baselines that include masking or blurring out entities by having access to ground truth mask annotations for people in the images. We also propose a baseline that simply adds noise to intermediate representations -- thus reducing the ability to predict gender from features, but often at a significant compromise in task accuracy. 
Of all methods considered, only adversarial debiasing provided a better trade-off compared to randomizing model predictions, and we were able to reduce bias amplification by 53-67\% while only sacrificing 1.2 - 2.2 points in accuracy.

\label{sec:intro}

\section{Related Work}
\vspace{-0.05in}
Recently, researchers have demonstrated that machine learning models tend to replicate societal biases present in training datasets. 
Concerns have been raised for applications such as recommender systems~\cite{yao2017beyond}, credit score prediction~\cite{hardt2016equality}, online news~\cite{ross2011women}, and others~\cite{kay2015unequal} and in response various approaches have been proposed to mitigate bias~\cite{agarwal2018reductions,pmlr-v80-hashimoto18a}.  
However, most previous work deals with issues of resource allocation~\cite{dwork2012fairness,feldman2015certifying} where the focus is on calibrating predictions.
Furthermore, works in this domain often assume protected variables are explicitly specified as features, making the goal of calibration more clearly defined. However in visual recognition, representations for protected attributes are automatically inferred from raw data.

There are also works addressing biases in images~\cite{ryu2017,zhao2017men,stock2017convnets,pmlr-v81-buolamwini18a,misra2016seeing,burns2018women}. 
Zhao~et~al~\cite{zhao2017men} reduces bias in structured prediction models where gender is one of the target variables. Burns~et~al~\cite{burns2018women} attempts to calibrate gender predictions of a captioning system by modifying the input image. In contrast, our work focuses on models that are not aimed at predicting gender, which is a more common scenario. Calibration methods would not be effective to debias in our proposed setup, as gender is not one of the outputs.

Our work is motivated by previous efforts on adversarial debiasing in various other tasks and domains~\cite{zhang2018mitigating,beutel2017data,xie2017controllable,elazar2018adversarial,ZZLWC18,ganin2016domain}. 
We provide further details about this family of methods in the body of the paper, and adopt this framework for debiasing the intermediate results of deep neural networks. 
Our work advances the understanding of this area by exploring what parts of deep representations are the most effective to debias under this approach, and we are the first to propose a way to visualize such debiased representations.

Issues of dataset bias have been addressed in the past the computer vision community~\cite{torralba2011unbiased,khosla2012undoing,tommasi2017deeper}. Torralba and Efros~\cite{torralba2011unbiased} showed that it was possible to identify the source dataset given image samples for a wide range of standard datasets, and~\cite{khosla2012undoing} addresses this issue by learning shared parameters across datasets. More recently, Tommasi~et~al~\cite{tommasi2017deeper} provided a fresher perspective on this issue using deep learning models. There are strong connections with these prior works when dataset source is to be taken as a protected variable. Our notion of bias is more closely related to the notion of bias used in the fairness in machine learning literature, where there is protected variable (e.g.~gender) for which we want to learn unbiased representations (e.g.~\cite{zemel2013learning}). 

In terms of evaluation, researchers have proposed different measurements for quantifying fairness~\cite{hardt2016equality,kusner2017counterfactual,dwork2012fairness}. In contrast, we try to reduce bias in the feature space. We adopt and further develop the idea of \emph{leakage} as an evaluation criteria, as proposed by Elazar and Goldberg~\cite{elazar2018adversarial} to debias text representations. 
We significantly expand the \emph{leakage} formulation and propose \emph{dataset leakage}, and \emph{model leakage} as measures of bias in learned representations.

Building models under fairness objectives is also related to feature disentangling methods~\cite{tenenbaum2000separating,rifai2012disentangling,liu2014feature,liu2017adaptive,liu2018exploring}. However, most research in this domain has focused on facial analysis -- where there is generally more well aligned features. This general area of work is also related to efforts in building privacy preserving methods~\cite{upmanyu2009efficient,sokolic2017learning,Wu_2018_ECCV,kim2019training}, where the objective is to obfuscate the input while still being able to perform a recognition task. In contrast, in fairness methods, there is no requirement to obfuscate the inputs, and in particular the method proposed in this paper is most effective when applied to intermediate feature representations. 

\label{sec:related}

\section{Leakage and Amplification}
Many problems in computer vision inadvertently reveal demographic information in images.
For example, in COCO, images of plates contain significantly more women than men. If a model predicts that a plate is in the image, we can infer there is likely a woman as well.
We refer to this notion as \emph{\it leakage}.
In this section, we present formal definitions of leakage for a dataset and models, and a measure for quantifying bias amplification as summarized in Figure~\ref{fig:lead}.

\vspace{0.04in}
\noindent{\bf Dataset Leakage: }
Given an annotated dataset $\mathcal{D}$ containing instances $(X_i, Y_i, g_i)$, where $X_i$ is an image annotated with a set of task-specific labels $Y_i$ (e.g., objects), and a protected attribute $g_i$ (e.g., the image contains a male/female person)\footnote{In this paper, we assume gender as binary due to the available annotations, but the work could be extended to non-binary, as well as a broader set of protected attributes, such as race or age.},  
we say that a particular annotation $Y_i$ leaks information about $g_i$ if there exists a function $f$ such that $g_i \approx f(Y_i)$.
We refer to this $f$ as an \emph{attacker} as it tries to reverse engineer information about protected attributes in the input image $X_i$ only from its task-specific labels $Y_i$.
To measure leakage across a dataset, we train such an attacker and evaluate it on held out data.
The performance of the attacker, the fraction of instances in $\mathcal{D}$ that leak information about $g_i$ through $Y_i$, yields an estimate of leakage:
\begin{equation*}
\label{eq:dataset-leakage}
\lambda_\mathcal{D} = \frac{1}{|\mathcal{D}|} \sum\nolimits_{(Y_i,g_i) \in \mathcal{D}} \mathbb{1}[f(Y_i) == g_i],
\end{equation*}
where $\mathbb{1}[\cdot]$ is the indicator function. 
We extend this definition of leakage to assess how much gender is revealed at different levels of accuracy, where errors are due entirely to chance.
We define dataset leakage at a performance $a$ by perturbing ground truth labels, with some function $r(Y_i,a)$, such that the overall accuracy of the changed labels with respect to the ground truth achieves an accuracy $a$:
\begin{equation*}
\lambda_D(a) = \frac{1}{|\mathcal{D}|} \sum\nolimits_{(Y_i,g_i) \in \mathcal{D}} \mathbb{1}[f(r(Y_i,a)) == g_i],
\end{equation*}

This allows us to measure the leakage of a model whose performance is $a$ and whose mistakes cannot be attributed to systematic bias.
Across all experiments, we use F1 as the performance measure, and $\lambda_D = \lambda_D (1.0)$, by definition.

\vspace{0.04in}
\noindent{\bf Model Leakage: }
Similar to dataset leakage, we would like to measure the degree a model, $M$ produces predictions, $\hat{Y}_i = M(X_i)$, that leak information about the protected variable $g_i$.
We define model leakage as the percentage of examples in $\mathcal{D}$ that leak information about $g_i$ through $\hat{Y}_i$.
To measure prediction leakage, we train a different attacker on $\hat{Y}_i$ to extract information about $g_i$:
\begin{equation*}
\label{eq:model_leakage}
\lambda_M(a) = \frac{1}{|\mathcal{D}|} \sum\nolimits_{(\hat{Y}_i,g_i) \in \mathcal{D}} \mathbb{1}[f(\hat{Y}_i) == g_i)],
\end{equation*}
where $f$ is a attacker function trained to predict gender from the outputs of model $M$ which has an accuracy score $a$.
\vspace{0.04in}
\noindent{\bf Bias Amplification: } Formally, we define the bias amplification of a model $p$, as the difference between the \emph{model leakage} and the \emph{dataset leakage} at the same accuracy $a$.
\begin{equation}
\label{eq:bias_amplification}
\Delta = \lambda_M(a) - \lambda_D(a)
\end{equation}

\noindent Note that $\lambda_D(a)$ measures the leakage of an ideal model which achieves a performance level $a$ but only makes mistakes randomly, not due to systematic bias.
A model with $\Delta$ larger than zero leaks more information about gender than we would expect even from simply accomplishing the task defined by the dataset.
This represents a type of amplification on the reliance on protected attributes to accomplish the prediction task.
In Eq.~\eqref{eq:bias_amplification}, $a$ could be any performance measurement but we use F1 score throughout our experiments.
We show later in Section~\ref{sec:leakage} that all models we evaluated leak more information than we would expect and even leak information when the dataset does not.

\vspace{0.02in}
\noindent{\bf Creating an Attacker: }
Ideally, the attacker should be a Bayes optimal classifier, which makes the best possible prediction of $g$ using $Y$. However, 
in practice, we need to train a model to do prediction for every model, and we use a deep neural network to do so. 
Yet, we are not guaranteed to obtain the best possible function for mapping $y$ to $g$. Thus, it is important to consider the reported leakage as a lower bound on true leakage.
In practice, we find that we can robustly estimate $f$ (see Section ~\ref{sec:leakage}: Attacker Learning is Robust).

\label{sec:general setup}

\section{Bias Analysis}
\label{sec:leakage}
\begin{table*}[t]
\centering
\small
\begin{tabular}{|l|l|c|c||c|c|c||c|c|c|}
\hline
             & \multicolumn{3}{c||}{Statistics} & \multicolumn{3}{c||}{Leakage} & 
             \multicolumn{3}{c|}{Performance} \\
             dataset& split& \#men & \#women & $\lambda_D$ & $\lambda_M(\text{F1})$ & $\lambda_D(\text{F1})$ & $\Delta$ & mAP & F1 \\ 
             \hline
\multirow{4}{*}{COCO~\cite{lin2014microsoft}}&original CRF &   $16,225$ &$6,601$ &$67.72\pm0.31$ & $73.20\pm0.59$ & $60.35$ & $12.85$ &$57.77$& $52.52$ \\ 
&no gender &   $16,225$ &$6,601$    &$67.72\pm0.31$ & $70.46\pm0.36$ & $60.53$ & $9.93$ & $58.23$ & $53.75$\\
&($\alpha=3$)                                           &  $10,876$ & $6,598$   &$62.00\pm0.98$ & $67.78\pm0.29$ & $57.50$ & $10.28$ & $57.04$ & $52.60$\\ 
&($\alpha=2$)                                           &  $8,885$ &$6,588$     &$56.77\pm1.45$ & $64.45\pm0.56$ & $54.72$ & $9.73$& $56.21$ & $51.95$\\ 
&($\alpha=1$)                                           &  $3,078$ & $3,078$    &$53.15\pm1.10$ & $63.22\pm1.11$ & $52.85$ & $10.37$& $48.23$ & $42.89$  \\ \hline
\multirow{4}{*}{imSitu~\cite{yatskar2016}}&original CRF    &  $14,199$& $10,102$   &$68.26\pm0.31$ & $78.43\pm0.26$  & $56.58$ & $21.85$ & $41.83$ &$40.75$      \\ 
&no gender     &  $14,199$& $10,102$   &$68.26\pm0.31$ & $76.93\pm0.20$ & $56.46$ & $20.47$ & $41.02$ & $40.11$      \\ 
&($\alpha=3$)                                           &  $11,613$ & $9,530$   &$68.11\pm0.55$ & $75.79\pm0.49$ & $55.98$ & $19.81$ & $39.20$ & $37.64$  \\ 
&($\alpha=2$)                                           &  $10,265$ & $8,884$   &$68.15\pm0.32$ & $75.46\pm0.32$ & $55.74$ & $19.72$ & $37.53$ & $36.41$ \\ 
&($\alpha=1$)                                           &  $7,342$ & $7,342$    &$53.99\pm0.69$ & $74.83\pm0.34$ & $53.20$ & $21.63$ & $34.63$ & $33.94$ \\ \hline
\end{tabular}
\vspace{0.05in}
\caption{In this table we show for different splits in COCO and imSitu, (1) {$\lambda_D$}, dataset leakage or the accuracy obtained by predicting gender from ground truth annotations, showing that our data balancing approach successfully achieves significantly reducing this type of leakage (2) {$\lambda_M(\text{F1})$}, model leakage or the accuracy obtained by a model trained to predict gender on the outputs of a model trained on the target task, the last two columns show the mAP and F1 score of the model, and (3) {$\lambda_D(\text{F1})$}, dataset leakage at a certain performance leverl, or the leakage of a model with access to ground truth annotations but with added noise so that its accuracy matches that of a model trained on this data, i.e. same F1 as shown in the last column. (4) {$\Delta$}, bias amplification, the difference between model leakage and dataset leakage at the same performance level, indicating how much more leakage the model is exhibiting over chance.}
\label{tab:natural_bias}
\end{table*}

In this section we summarize our findings that both imSitu and COCO leak information about gender.
We show that models trained on these datasets leak more information than would be expected (1) when models are required to predict gender through a structured predictor that jointly predicts labels and gender, (2) when models are required to predict only labels, and (3) even when not predicting gender and datasets were balanced such that each gender co-occurs equally with target labels.
Table~\ref{tab:natural_bias} summarizes our results.
\subsection{Experiment Setup}
We consider two tasks: (1) multi-label classification in the COCO dataset~\cite{lin2014microsoft}, including the prediction of gender,
and (2) imSitu activity recognition, a multi-classification task for people related activities.

\vspace{0.02in}
\noindent {\bf Datasets: } 
We follow the setup of existing work for studying bias in COCO and imSitu~\cite{zhao2017men}, deriving gender labels from captions and ``agent'' annotations respectively.
For the purpose of analysis, we exclude ``person'' category and only use images containing people. 
We have $22826$, $5367$, $5473$ and $24301$, $7730$, $7669$ images in the training, validation and testing set for COCO and imSitu respectively.

\vspace{0.02in}
\noindent {\bf Models: }
For both object and activity recognition, we use a standard ResNet-50 pretrained on Imagenet (ILSVRC) as the underlying model by replacing the last linear layer. 
We also consider the Conditional Random Field (CRF) based model in~\cite{zhao2017men} when predicting gender jointly with target variables.
Attackers are a 4-layer multi-layer perceptron (MLP) with BatchNorm and LeakyReLU in between.

\vspace{0.02in}
\noindent{\bf Metrics: }
We use mAP, or the mean across categories of the area under the precision-recall curve, and F1 score for both tasks by using the discrete output predictions of the model.

\vspace{0.02in}
\noindent {\bf Computing Leakage: }
Model leakage is predicted from pre-activation logits while dataset leakage is predicted from binary labels.
Attackers are trained and evaluated with an equal amount images of men and women.

\vspace{0.02in}
\noindent{\bf Training Details: }
All models are developed and evaluated on the same dev and test sets from the original split. 
We optimize using Adam~\cite{Adam} with a learning rate of $10^{-4}$ and a minibatch size of $32$ to train the linear layers for classification.
We then fine-tune the model with a learning rate of $5\times 10^{-6}$. 
We train all attackers for $100$ epochs with a learning rate of $5\times 10^{-5}$ and a batch size of $128$, keeping the snapshot that performs best on the dev set.

\subsection{Results}
\label{sec:data-splits}
\noindent{\bf Dataset Leakage: }
Dataset leakage measures the degree to which ground truth labels can be used to estimate gender. 
The rows corresponding to ``original CRF'' in Table~\ref{tab:natural_bias} summarize dataset leakage in imSitu and COCO ($\lambda_D$).
Both datasets leak information: the gender of a main entity in the image is extractable from ground truth annotations 67.72\%  and 68.26\% for COCO and imSitu, respectively.

\begin{figure}[t]
    \centering
      \vspace{-5pt}
      \includegraphics[width=.4\textwidth]{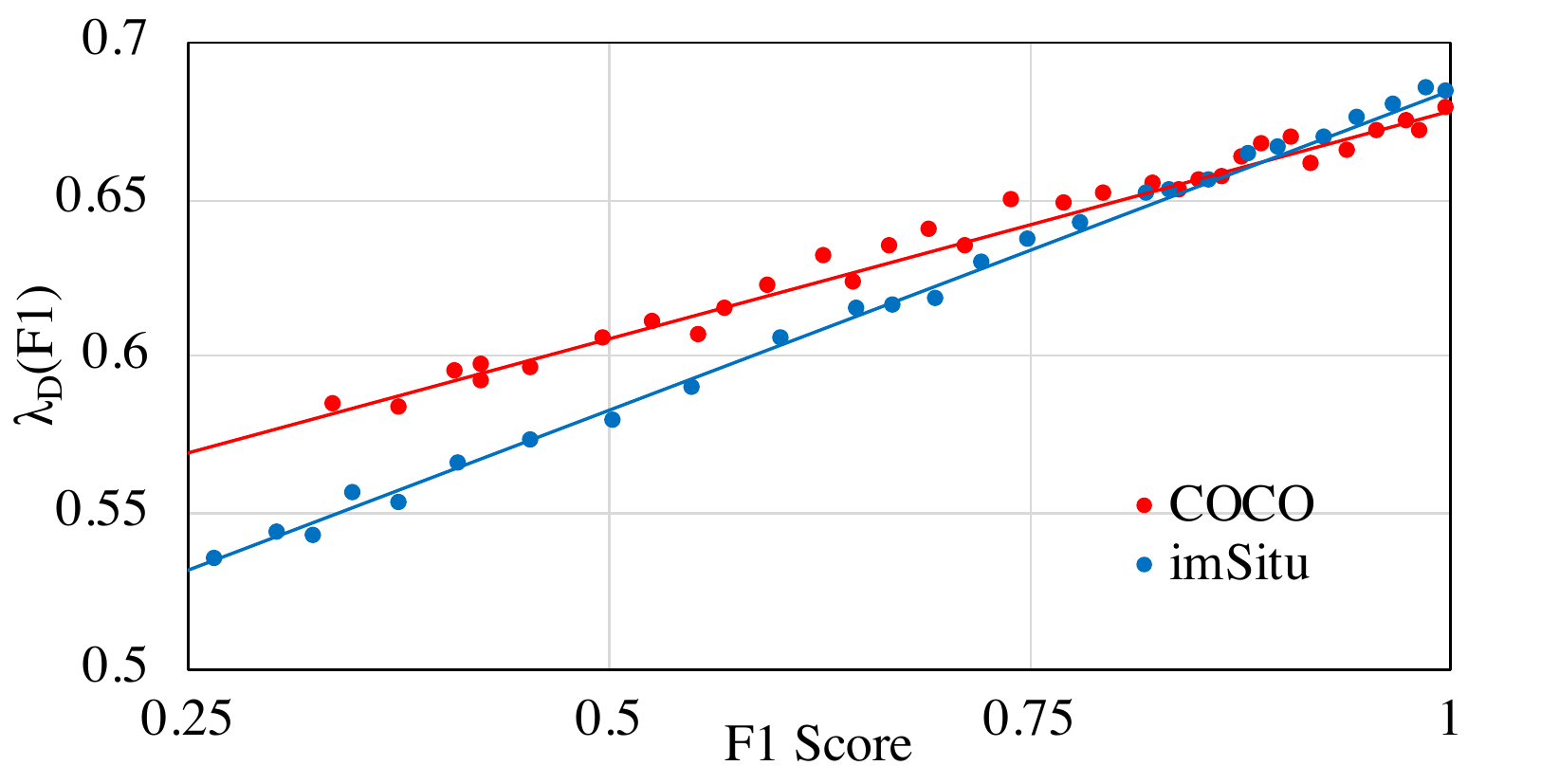}
    \caption{Dataset leakage in COCO and imSitu as function of F1 score. Ground truth labels were randomly flipped to simulate a method that performs at different levels of F1 score. We refer to this accuracy adjusted leakage as \emph{$\lambda_D(F1))$}, or the amount we would expect a method to leak given its performance level.}
    \label{fig:natura_bias}
    \vspace{-5pt}
\end{figure}

\vspace{0.04in}
\noindent{\bf Bias Amplification: }
Bias amplification ($\Delta$) captures how much more information is leaked than what we expect from a similar model which makes mistakes entirely due to chance.
Dataset leakage needs to be calibrated with respect to model performance for computing bias amplification.
To do so, we randomly flip ground truth labels to reach various levels of accuracy.
Figure~\ref{fig:natura_bias} shows dataset leakage at different performance levels in COCO and imSitu.
The relationship between F1 and leakage is roughly linear.
In Table~\ref{tab:natural_bias}, we report adjusted leakage for models at appropriate levels ($\lambda_D(\text{F1})$).
Finally, bias amplification ($\Delta$) 
can be computed by taking the difference between adjusted dataset leakage ($\lambda_D(\text{F1})$) and model leakage ($\lambda_M(\text{F1}))$.

Models trained on standard splits of both COCO and imSitu that jointly predict gender and target labels (the original rows in Table~\ref{tab:natural_bias}), all leak significantly more gender information than we would expect by chance.
Surprisingly, imSitu is more gender balanced than COCO but actually leaks significantly more information than models trained on COCO.
When models are no longer required to predict gender, they leak less information than before but still more than we would expect (the \emph{no gender} rows in Table~\ref{tab:natural_bias}).

\vspace{0.04in}
\noindent{\bf Alternative Data Splits: }
It is possible to construct datasets which leak less through subsampling. 
We obtain splits more balanced in male and female co-occurrences with labels by imposing the constraint that neither gender occurs more frequently with any output label by a ratio greater than $\alpha$: 
\vspace{-5pt}
\begin{equation} 
\forall y :  1/\alpha < \# (m, y)  / \# (w, y) < \alpha,
\vspace{-5pt}
\end{equation}
where $\# (m, y)$ and $\# (w, y)$ are the number of occurrences of \emph{men} with label $y$ and of \emph{women} with label $y$ respectively. 
Enforcing this constraint in imSitu is trivial because each image is only annotated with one verb: we simply sample the over-represented gender to pass the above constraints. 
For COCO, we heuristically enforce this constraint since each image contains multiple object annotations. 
We try to make every object satisfy this constraint one at a time, removing images having less objects. 
We iterate through all objects until this process converges and all objects satisfy the constraint. 
We create splits for $\alpha \in \{3, 2, 1\}$.\footnote{Practically satisfying $\alpha = 1$ is in-feasible, but our heuristic is able to find a set where $\alpha = 1.08$.}

Table~\ref{tab:natural_bias} rows $\alpha=\{3,2,1\}$ summarize results for rebalancing data with respect to gender. 
As we expect, decreasing values of $\alpha$ yields smaller datasets with less dataset leakage but worse predictors because there is less data.
Yet model leakage does not reduce as quickly as dataset leakage, resulting in nearly no change in bias amplification. 
In fact, when there is nearly no dataset leakage, models still leak information.
Likely this is because it is impossible to balance \emph{unlabeled} co-occurring features with gender (e.g. COCO only has annotations for $80$ objects) and the models still rely on these features to make predictions. In summary, {\bf balancing the co-occurance of gender and target labels does not reduce bias amplification in a meaningful way.}

\begin{table}[t]
\small
\centering
\begin{tabular}{|c|c|}
\hline
{Attacker} & $\lambda_M$ \\ 
\hline
1 layer , ---------- , all data &  $68.82\pm0.35$ \\
2 layer , 100 dim , all data &  $70.83\pm0.58$\\
2 layer , 300 dim , all data & $71.03\pm0.52$\\
4 layer , 300 dim , all data & $70.46\pm0.36$ \\ 
4 layer , 300 dim , 75\% data & $69.93\pm0.51$  \\
4 layer , 300 dim , 50\% data & $69.89\pm0.98$\\
4 layer , 300 dim , 25\% data & $68.54\pm1.10$  \\
\hline
\end{tabular}
\vspace{0.1in}
\caption{ Varying attacker architecture and training data when estimating model leakage on the original COCO. The leakage estimate is robust to significant changes, showing that estimation of leakage with our adversaries is largely easy and stable. }
\label{tab:robust_attacker}
\vspace{-0.2in}
\end{table}
\vspace{0.05in}
\noindent{\bf Attacker Learning is Robust: }
Measuring leakage relies on being able to consistently estimate an attacker. 
To verify that leakage estimates are robust to different architectures and data settings on the attacker side, we conduct an ablation study in Table~\ref{tab:robust_attacker}. 
We vary the attacker architecture and the amount of training data to measure model leakage ($\lambda_M$).
Except an attacker with 1-layer, none of the others vary in their estimation of leakage by more than 2 points.

\section{Adversarial Debiasing}
\label{sec:debiasing}
In this section we show the effectiveness of a method for reducing leakage through training with an auxiliary adversarial loss which effectively removes gender information from intermediate representations. We additionally propose a way to visualize the effects of this approach on the input space, to inspect the type of information being removed.

\subsection{Method Overview}
\label{sec:method_adv}
We propose a simple formulation for reducing the amount of leakage in a model, summarized in Figure~\ref{fig:method}.
We hypothesize that models leak extra information about protected attributes because the underlying representation is overly sensitive to features related to those attributes.
As such, we encourage models to build representations from which protected attributes can not be predicted.

Our methods rely on the construction of a \emph{critic}, $c$, which attempts to predict protected information from an intermediate representation, $h_i$ for a given image $X_i$, of a predictor, $p$. 
The critic attempts to minimize a loss over the amount of information it can extract:
\begin{equation*}
\sum\nolimits_{(h_i,g_i) \in \mathcal{D}}L_c( c(h_i) , g_i),
\end{equation*}
while the predictor tries to minimize its loss over the task specific predictions while increasing the critic's loss:
\begin{equation*}
L_p = \sum\nolimits_{(X_i,h_i, Y_i) \in \mathcal{D}} [L( p(X_i) , Y_i) - \lambda L_c( c(h_i) , g_i)].
\end{equation*}
In both cases, $L$ is the cross-entropy loss, and when optimizing $L_p$ we do not update $c$, and trade-off task performance with sensitivity to protected attributes with $\lambda$.

We also experiment with optimizing the adversarial loss on the input feature space by leveraging an encoder-decoder model that auto-encodes the input image $X_i$.  In order to accomplish this goal, we add an additional loss with a weight parameter $\beta$ to the predictor as follows:
\begin{figure*}[t]
    \centering
      \includegraphics[width=.9\textwidth]{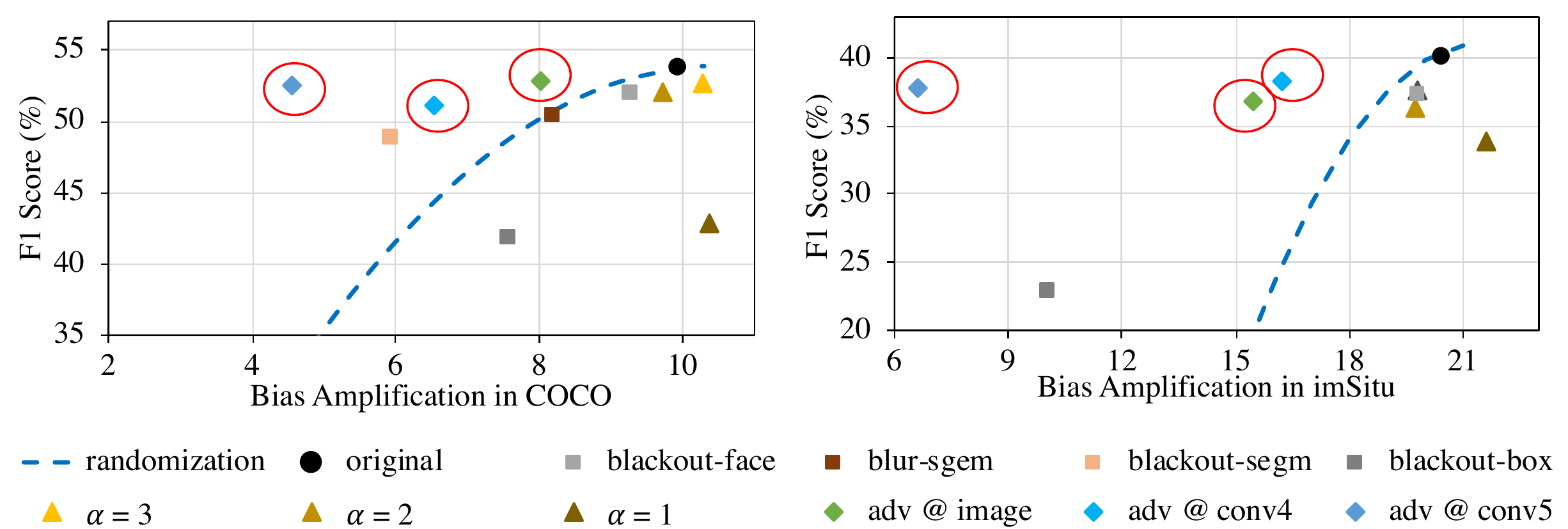}
    \caption{Bias amplification as a function of F1 score on COCO object classification and imSitu action recognition. Models in the top left have low leakage and high F1 score. The blue dashed line indicates bias and performance of adding progressively more noise to the original model representation.  Our adversarial methods (circled) are the ones which make a better trade-off between performance and bias amplification than randomization and other baselines. }
    \label{fig:coco_imSitu}
\end{figure*}
$$
L_p = \sum\nolimits_{i} \left[ \beta|X_i - \hat{X}_i|_{\ell_1}
\!+\! L( p(\hat{X}_i) , Y_i) 
\!-\! \lambda L_c( c(\hat{X}_i) , g_i)\right]
$$

Where $\hat{X}_i = {M}_i \cdot X_i$, which is the original image element-wise multiplied with a mask $M_i$ generated by an encoder-decoder bottleneck network with input $X_i$. So the first term is encouraging the mask to maintain the information in the original image, the second term is trying to obtain correct task-specific predictions from the masked input, and the third term is adversarially trying to obscure gender by modifying the mask. This is similar to the proposed experiment in Palacio~et~al~\cite{palacio2018deep} where instead, the outputs of an autoencoder are directly fed to a convolutional neural network trained to recognize objects in order to interpret the patterns learned by the network. In contrast, our objective is to visualize what the adversary learned to obfuscate while trying to preserve accurate results.

\subsection{Implementation Details}
We first train the classification layers (linear classifiers) with $10^{-4}$ as learning rate and a batch size of $32$ until the performance plateaus. We then incorporate the adversarial loss, and fine-tune the model end-to-end using a learning rate $5\!\times\!10^{-6}$.
Before activating the adversarial loss, we first train the gender classification branch so that its gradients provide useful guidance for feature removal during adversarial training. In every batch, we sample the same amount of male and female images for training this adversary. 
\subsection{Models}
\paragraph{Adversarial Methods} We consider three different types of adversaries which try to remove leakage at different stages in a ResNet-50 classification network.
\begin{itemize}[noitemsep,topsep=1pt,parsep=3pt,partopsep=0pt,leftmargin=*]
    \setlength\itemsep{0pt}
    \item {\bf adv @ image}, or removing gender information directly at the image. We use  U-Net~\cite{ronneberger2015u} as our encoder-decoder network to predict a mask $M_i$. The original image is point-wise multiplied with this mask and then fed to two branches. The first branch is a ResNet-18 which attempts to detect gender (the adversary) and the second branch is a ResNet-50 for classifying the target categories. 
    \item {\bf adv @ conv4}, removes gender information from an intermediate hidden representation of ResNet-50 (on the 4th convolutional block). We use an adversary with 3 convolutional layers and 4 linear layers. 
    \item {\bf adv @ conv5}, removes gender information from the final convolutional layer of ResNet-50. We use a linear adversary which takes as input a vectorized form of the output feature map and uses a 4-layer MLP for classification.  
\end{itemize}

\noindent{\bf Baselines: } We consider several alternatives to adversarial training to reduce leakage, including some that have access to face detectors and ground truth segment annotations. 
\begin{itemize}[noitemsep,topsep=1pt,parsep=3pt,partopsep=0pt,leftmargin=*]
    \setlength\itemsep{0pt}
    \item {\bf Original}: The basic recognition model, trained on the original data, without any debiasing attempt.
    \item {\bf Randomization}: Adding Gaussian noise at increasing magnitudes to the pre-classification embedding layer of the original model. We expect larger noise to reduce more leakage while preventing the model from effectively classifying images.
    \item {\bf Alternative Datasets}: We also consider constructing alternative data splits through downsampling approaches that reduce dataset leakage. We refer to this alternative data splits as $\alpha = {1,2,3}$, as defined in section~\ref{sec:data-splits}. 
    \item {\bf Blur}: Consists of blurring people in images when ground truth segments are available (COCO only).
    \item {\bf Blackout - Face}: Consists of blacking out the faces in the images using a face detector. 
    \item {\bf Blackout - Segm}: Consists of blacking out people in images when ground truth segments are available (COCO only). This aggressively removes features such as skin and clothing. It may also obscure objects with which people are closely interacting with.
    \item {\bf Blackout - Box}: Consists of blacking out people using 
    ground truth bounding boxes (COCO and imSitu). This removes large regions of the image around people, likely removing many objects and body pose cues. 
\end{itemize}

\begin{table}[t]
\small
\centering
\setlength\tabcolsep{4pt}
\begin{tabular}{|l||c|c||c|c|c|}
\hline
& \multicolumn{2}{c||}{Leakage} & \multicolumn{3}{c|}{Performance} \\
              & $\lambda_M(\text{F1})$ & $\lambda_D(\text{F1})$ & $\Delta$ & mAP & F1 \\ 
                 \hline
original CRF & $73.20$ & $60.35$ & $12.85$ & $57.77$ & $52.53$ \\
CRF + RBA &$73.31$&$60.16$& $13.15$ & $56.46$& $51.28$ \\
CRF + adv &$65.00$&$60.19$& $\bm{4.81}$ & $56.68$& $51.48$ \\
\hline
\end{tabular}
\vspace{0.10in}
\caption{Model Leakage and performance trade-offs for RBA (Reducing Bias Amplification, proposed in~\cite{zhao2017men}) and our adversarial training methods. We adopt the CRF based model~\cite{zhao2017men} to predict COCO objects as well as the gender. Our method reduces more than $60\%$ bias amplification while RBA fails to do so.}
\label{tab:CRF}
\vspace{-0.1in}
\end{table}

\begin{table}[t]
\small

\centering
\setlength\tabcolsep{2.5pt}
\begin{tabular}{|l||c|c||c|c|c|}
\hline
& \multicolumn{2}{c||}{Leakage} & \multicolumn{3}{c|}{Performance} \\
              & $\lambda_M(\text{F1})$ & $\lambda_D(\text{F1})$ & $\Delta$ & mAP & F1 \\ 
                 \hline
original & $70.46$ & $60.53$ & $9.93$ & $\bm{58.23}$ & $\bm{53.75}$ \\
blackout-face &$69.53$&$60.24$& $9.29$ & $55.93$& $51.81$ \\
blur-segm & $68.19$ & $59.99$ & $8.20$ & $55.06$ & $50.26$ \\
blackout-segm &$65.72$&$59.76$& $5.96$ & $53.78$& $48.72$ \\
blackout-box & $64.00$ &$58.71$& $5.29$ & $47.42$ & $41.81$ \\
adv @ image &$68.49$& $60.47$ & $8.02$ & $56.14$ & $52.82$ \\
adv @ conv4 &$66.66$&$60.12$& $6.54$&$55.18$ & $51.08$ \\
adv @ conv5 &$64.92$&$60.35$& $\bm{4.57}$ & $56.35$& $52.54$ \\
\hline\hline
($\alpha=1$) & $63.22$ & $52.85$ & $10.37$& $48.23$ & $42.89$  \\
adv @ conv5 &$\bm{54.91}$&$52.40$& $\bm{2.51}$ & $43.71$& $38.98$ \\
\hline
\end{tabular}
\vspace{0.10in}
\caption{Model leakage and performance trade-offs for different baselines (rows 1-5) and our adversarial training methods (rows 6-8) on COCO object classification. Our methods make significantly better trade-offs than baselines, even improving on methods which use ground truth detection and segmentation. Applying adversarial training on balanced dataset reaches lowest model leakage ($54.91$) and bias amplification ($2.51$). }
\label{tab:coco_debias}
\vspace{-0.1in}
\end{table}

\begin{table}[t]
\small
\centering
\setlength\tabcolsep{2.5pt}
\begin{tabular}{|l||c|c||c|c|c|}
\hline
 & \multicolumn{2}{c||}{Leakage} & \multicolumn{3}{c|}{Performance} \\
   & $\lambda_M(\text{F1})$ & $\lambda_D(\text{F1})$ & $\Delta$ & mAP & F1 \\ 
 \hline
original & $76.93$ & $56.46$ & $20.47$ & $\bm{41.02}$ & $\bm{40.11}$ \\
blackout-face &$75.69$&$55.91$& $19.78$ & $38.22$& $37.29$ \\
blackout-box & $63.14$ &$53.06$&$10.08$  & $21.76$ & $22.75$ \\
adv @ image & $71.32$ & $55.83$ &$15.49$ & $36.90$ & $36.88$ \\ 
adv @ conv4 & $72.39$ & $56.15$ & $16.24$ & $38.81$ & $38.35$ \\
adv @ conv5 &$62.65$ & $56.02$ & $\bm{6.63}$ & $38.91$& $37.85$ \\
\hline\hline
($\alpha=1$)& $74.83$ & $53.20$ & $21.63$ & $34.63$ & $33.94$ \\
adv @ conv5&$\bm{57.49}$ & $52.85$ & $\bm{4.64}$ & $30.78$& $30.37$ \\
\hline
\end{tabular}
\vspace{0.10in}
\caption{Model leakage and performance trade-offs for different baselines (rows 1-3) and our adversarial training methods (rows 4-6) on imSitu activity recognition. Our methods make significantly better trade-offs than baselines. Applying adversarial training on balanced dataset reaches lowest model leakage ($57.49$) and bias amplification ($4.64$). }
\label{tab:imSitu_debias}
\vspace{-0.1in}
\end{table}

\vspace{-0.2in}
\subsection{Quantitative Results}
\begin{figure}[t]
    \centering
      \includegraphics[width=.38\textwidth]{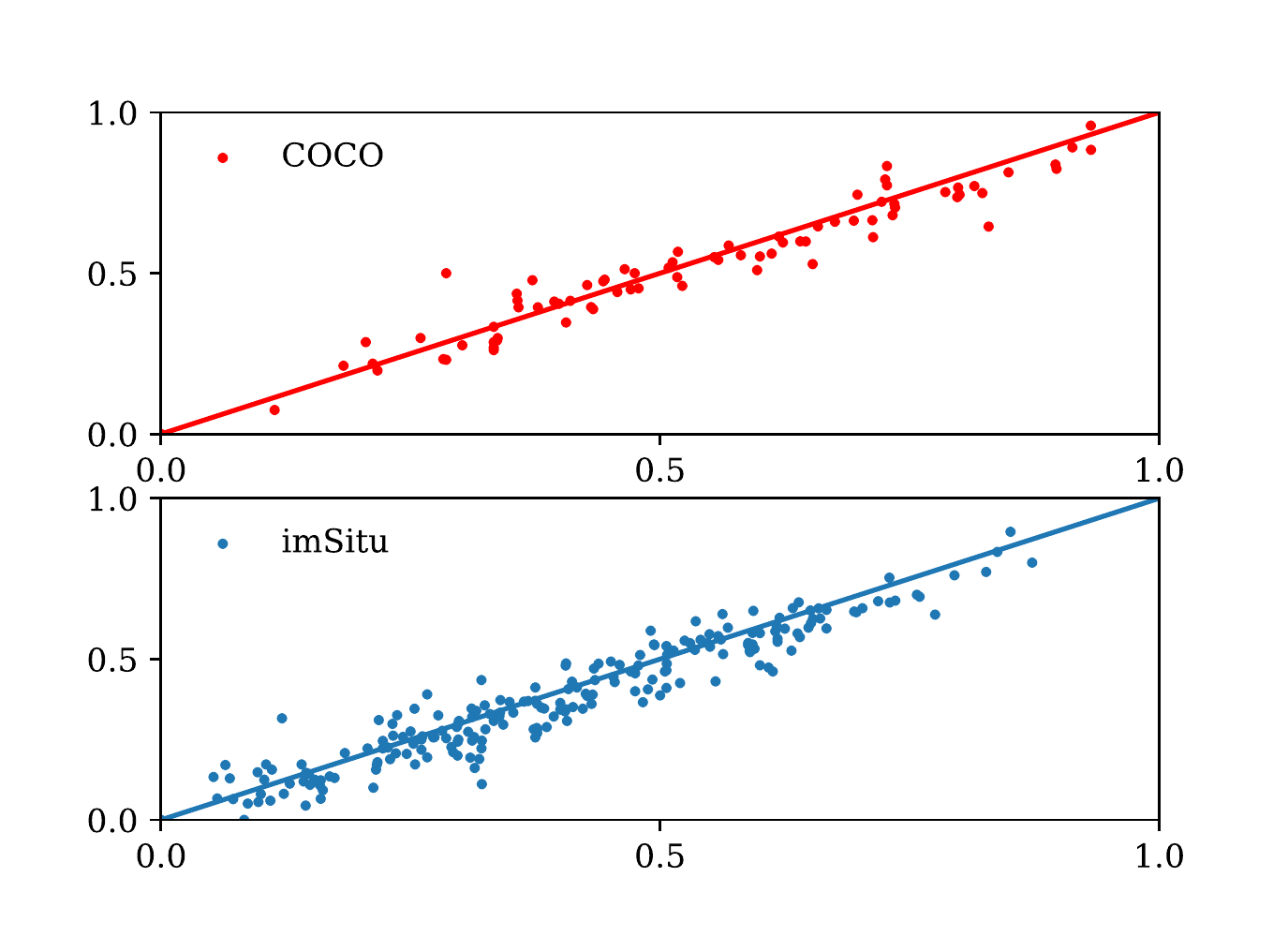}
    \caption{Performance change of every object/verb. X axis: F1 score before debiasing. Y axis: F1 score after debiasing. Across two datasets, most of objects/verbs are very close to the solid line ($y=x$), showing that no one class is disproportionately affected.}
    \label{fig:coco_imSitu_change}
\end{figure}

Table~\ref{tab:coco_debias} and Table~\ref{tab:imSitu_debias} summarize our results. 
Adversarially trained methods offer significantly better trade-offs between leakage and performance than any other method. We are able to reduce model leakage by over 53\% and 67\%  on COCO and imSitu respectively, while suffering only 1.21 and 2.26 F1 score degradation. 
Furthermore, no one class disproportionately suffers after our method (See Figure \ref{fig:coco_imSitu_change}).
We also compare our method with RBA~\cite{zhao2017men}, a debiasing algorithm proposed to maintain the similarity between the training data and model predictions. As shown in Table~\ref{tab:CRF}, the original CRF model predicts gender and objects, RBA fails to have reduce bias amplification. 
Figure~\ref{fig:coco_imSitu} further highlights that our methods are making extremely favorable trade-offs between leakage and performance, even when compared to methods that blur, black-out, or completely remove people from the images using ground truth segment annotations. 
Adversarial training is the only method that consistently improves upon simply adding noise to the model representation before prediction (the blue curves).

\begin{figure*}[t]
    \centering
      \includegraphics[width=0.95\textwidth]{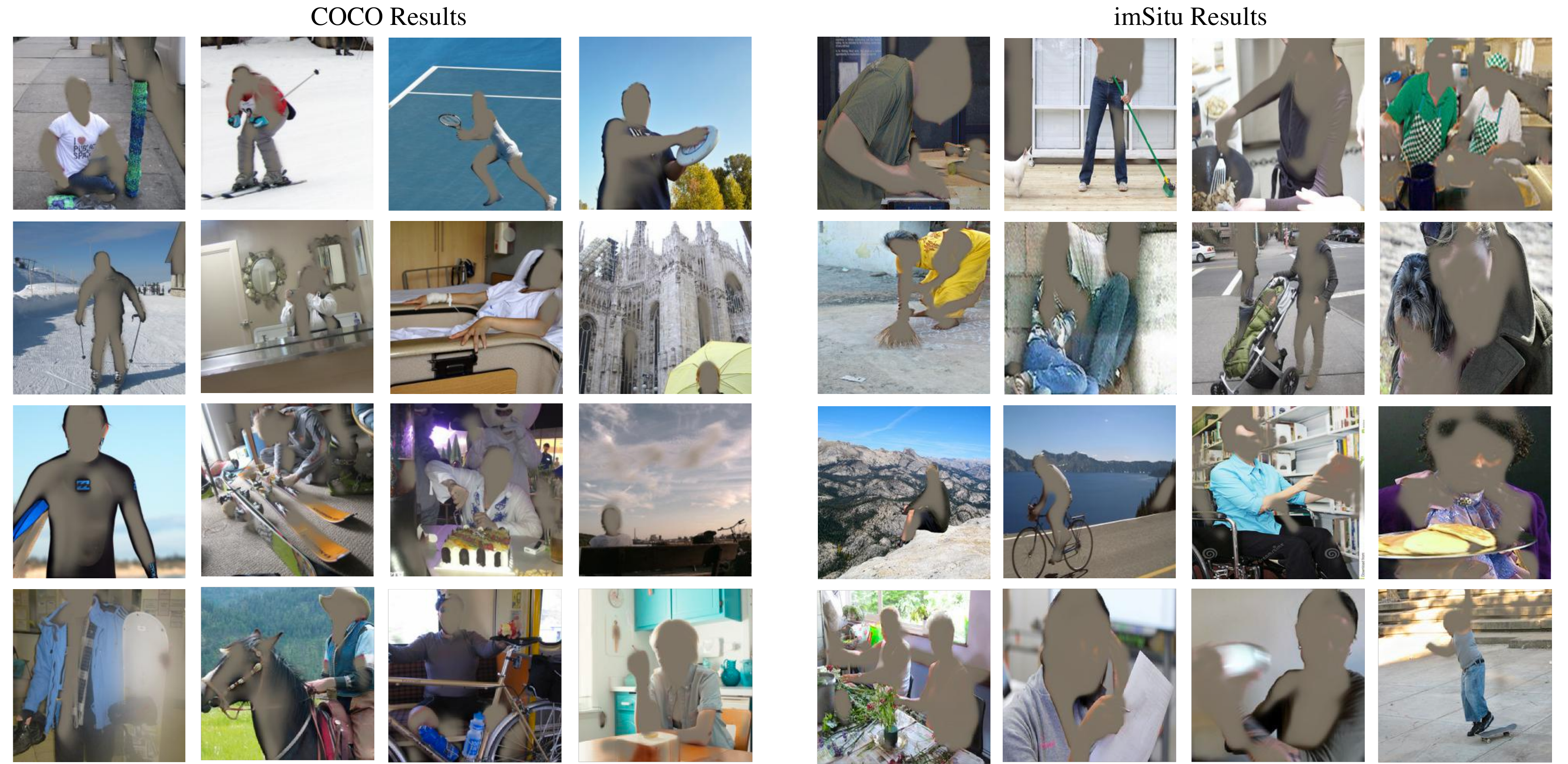}
    \caption{Images after adversarial removal of gender in image space by using a U-Net based autoencoder as inputs to the recognition model. While people are clearly being obscured from the image, the model selectively chooses to obscure only parts that would reveal gender such as faces but tries to keep information that is useful to recognize objects or verbs. 1st row: WWWM MMWW; 2nd row: MWWW WMWW; 3rd row: MMMW MMWM; 4th row: MMMW WWMM. W: woman; M: man.}
    \label{fig:qualitative-results}
\end{figure*}

\subsection{Qualitative Results}
  While adversarial removal works best when applied to representations in intermediate convolutional layers. In order to obtain interpretable results, we apply gender removal in the image space and show results in Fig.~\ref{fig:qualitative-results}. In some instances our method removes the entire person, in some instances only the face, in other cases clothing, and garments that might be strongly associated with gender. Our approach learns to selectively obscure pixels enough to make gender prediction hard but leaving sufficient information to predict other things, especially objects that need to be recognized such as \emph{frisbee}, \emph{bench}, \emph{ski}, as well as actions such as \emph{cooking}, \emph{biking}, etc. This is in contrast to our strong baselines that remove the entire person instances using ground-truth segmentation masks. A more sensible compromise is learned through the adversarial removal of gender without the need for segment-level supervision.

\section{Conclusion}
\label{sec:conclusion}
We introduced \emph{dataset leakage}, and \emph{model leakage} as measures of the encoded bias with respect to a protected variable in either datasets or trained models. We demonstrated that models amplify the biases in existing datasets for tasks that are not related to gender recognition. Moreover, we show that balanced datasets do not lead to unbiased predictions and that more fundamental changes in visual recognition models are nedeed. We also demonstrated an adversarial approach for the removal of features associated with a protected variable from the intermediate representations learned by a convolutional neural network. Our approach is superior to applying various forms of random perturbations in the representations, and to applying image manipulations that have access to significant privileged information such as people segments. We expect that the setup, methods, and results in this paper will be useful for further studies of representation bias in computer vision.

\vspace{0.08in}
\noindent {\bf Acknowledgements}
This research was supported partially by a Google Faculty Award, DARPA (HR0011-18-9-0019), and gift funding from SAP Research and Leidos Inc. We also acknowledge fruitful discussions with members of the Human-Machine Intelligence group through the Institute for the Humanities and Global Cultures at the University of Virginia.

{\small
\bibliographystyle{ieee_fullname}
\bibliography{egbib}
}

\end{document}